\documentclass[runningheads]{llncs}
\usepackage{graphicx}
\usepackage{mathtools}
\usepackage{amsmath}
\usepackage{url}
\usepackage{amssymb}
\usepackage[misc]{ifsym}
  
\begin{document}
\title{Face-Cap: Image Captioning using Facial Expression Analysis}
%
%
\author{Omid Mohamad Nezami\inst{1} \Letter \and
Mark Dras\inst{1} \and
Peter Anderson\inst{1,2} \and Len Hamey\inst{1}}
\authorrunning{Nezami et al.}
%
\institute{Department of Computing, Macquarie University, Sydney, Australia\\
\email{omid.mohamad-nezami@hdr.mq.edu.au}\\
\email{\{mark.dras,len.hamey\}@mq.edu.au}\\
\and
The Australian National University, Canberra, Australia\\
\email{peter.anderson@anu.edu.au}}
\maketitle              
\begin{abstract}
Image captioning is the process of generating a natural language description of an image. Most current image captioning models, however, do not take into account the emotional aspect of an image, which is very relevant to activities and interpersonal relationships represented therein. Towards developing a model that can produce human-like captions incorporating these, we use facial expression features extracted from images including human faces, with the aim of improving the descriptive ability of the model. In this work, we present two variants of our Face-Cap model, which embed facial expression features in different ways, to generate image captions. Using all standard evaluation metrics, our Face-Cap models outperform a state-of-the-art baseline model for generating image captions when applied to an image caption dataset extracted from the standard Flickr 30K dataset, consisting of around 11K images containing faces.  An analysis of the captions finds that, perhaps surprisingly, the improvement in caption quality appears to come not from the addition of adjectives linked to emotional aspects of the images, but from more variety in the actions described in the captions.

\end{abstract}
\section{Introduction}
\label{sec:intro}

Image captioning systems aim to describe the content of an image using computer vision and natural language processing. This is a challenging task in computer vision because we have to capture not only the objects but also their relations and the activities displayed in the image in order to generate a meaningful description. Most of the state-of-the-art methods, including deep neural networks, generate captions that reflect the factual aspects of an image~\cite{anderson2017bottom,fang2015captions,johnson2016densecap,karpathy2015deep,kiros2014unifying,vinyals2015show,xu2015show}; the emotional aspects which can provide richer and  attractive image captions are usually ignored in this process.
Emotional properties, including recognizing and expressing emotions, are required in designing intelligent systems to produce intelligent, adaptive, and effective results~\cite{lisetti1998affective}. Designing an image captioning system, which can recognize emotions and apply them to describe images, is still a challenge.

A few models have incorporated sentiment or other non-factual information into image captions~\cite{gan2017stylenet,mathews2016senticap,you2018image}; 
they typically require the collection of a supplementary dataset, with a sentiment vocabulary derived from that, drawing from work in Natural Language Processing \cite{pang-lee:2008} where sentiment is usually characterized as one of positive, neutral or negative. Mathews et al.~\cite{mathews2016senticap}, for instance, constructed a sentiment image-caption dataset via crowdsourcing, where annotators were asked to include either positive sentiment (e.g. \textit{a cuddly cat}) or negative sentiment (e.g. \textit{a sinister cat}) using a fixed vocabulary; their model was trained on both this and a standard set of factual captions. Gan et al.~\cite{gan2017stylenet} proposed a captioning model called StyleNet to add styles, which could include sentiments, to factual captions; they specified a predefined set of styles, such as humorous or romantic.

These kinds of models typically embody descriptions of an image that represent an \textit{observer's} sentiment towards the image (e.g. \textit{a cuddly cat} for a positive view of an image, versus \textit{a sinister cat} for a negative one); they do not aim to capture the emotional content of the image, as in Fig.~\ref{fig:example1}.  This distinction has been recognized in the sentiment analysis literature: the early work of \cite{pang-lee:2004:ACL}, for instance, proposed a graph-theoretical method for predicting sentiment expressed by a text's author by first removing text snippets that are positive or negative in terms of the actual content of the text (e.g. ``The protagonist tries to protect her good name'' as part of the description of a movie plot, where \textit{good} has positive sentiment) and leaving only the sentiment-bearing text that reflects the writer's subjective view (e.g. ``bold, imaginative, and impossible to resist''). We are interested in precisely this notion of content-related sentiment, in the context of an image.

In this paper, therefore, we introduce an image captioning model we term Face-Cap to incorporate emotional content from the images themselves: we automatically detect emotions from human faces, and apply the derived facial expression features in generating image captions. We introduce two variants of Face-Cap, which employ the features in different ways to generate the captions. The contributions of our work are:
\begin{enumerate}
	\item
	Face-Cap models that generate captions incorporating facial expression features and emotional content, using neither sentiment image-caption paired data nor sentiment caption data, which is difficult to collect.  To the authors' knowledge, this is the first study to apply facial expression analysis in image captioning tasks. 
	\item
	A set of experiments that demonstrate that these Face-Cap models outperform baseline, a state-of-the-art model, on all standard evaluation metrics.  An analysis of the generated captions suggests that they improve over baseline models by better describing the actions performed in the image.
	\item
	An image caption dataset that includes human faces which we have extracted from Flickr 30K dataset~\cite{young2014image}, which we term FlickrFace11K. It is publicly available\footnote{\url{https://github.com/omidmn/Face-Cap}} for facilitating future research in this domain.
\end{enumerate}

The rest of the paper is organized as follows. In Sec.\,\ref{sec:relwork}, related work in image captioning and facial expression recognition is described. In Sec.\,\ref{sec:models}, we explain our models to caption an image using facial expression analysis. To generate sentimentally human-like captions, we show how facial expression features are detected and applied in our image captioning models. Sec.\,\ref{sec:exper} presents our experimental setup
and the evaluation results. The paper concludes in Sec.\,\ref{sec:concl}.

\section{Related Work}
\label{sec:relwork}
In the following subsections, we review image captioning and facial expression recognition models as they are the key parts of our work.

\subsection{Image Captioning}
Recent image captioning models apply a CNN model to learn the image contents (encoding), followed by a LSTM to generate the image caption (decoding). This follows the paradigm employed in neural machine translation, using deep neural networks~\cite{sutskever2014sequence} to translate an image into a caption. In terms of encoding, they are divided into two categories: global encoding and fragment-level encoding~\cite{karpathy2016connecting}. The global approach encodes an image into a single feature vector, while the fragment-level one encodes the image fragments into separate feature vectors.

As a global encoding technique, Kiros et al.~\cite{kiros2014unifying} applied a CNN and a LSTM to capture the image and the caption information, separately. They made a joint multi-modal space to encode the information and a multi-modal log-bilinear model (in the form of a language model) to generate new captions. In comparison, Vinyals et al.~\cite{vinyals2015show} encoded image contents using a CNN and applied a LSTM to generate a caption for the image in an end-to-end neural network model. In general, the global encoding approaches generate captions according to the detected objects in an image; however, when the test samples are significantly different from the training ones in terms of the object locations and interactions, they often cannot generalize to the test samples in terms of appropriate captions.

\begin{figure}
	\centering
	\includegraphics[height=3.3 cm]{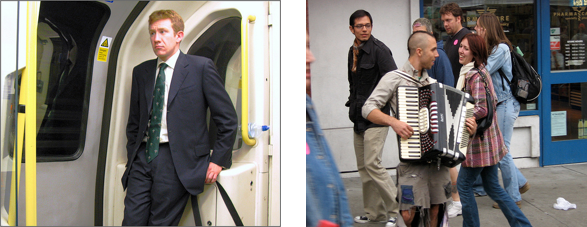}
	\caption{The examples of Flickr 30K dataset~\cite{young2014image} including sentiments. A man in a suit and tie with a \textbf{sad} look on his face (left) and a man on a sidewalk is playing the accordion while \textbf{happy} people pass by (right).}
	\label{fig:example1}
\end{figure}

With respect to fragment-level encoding, Fang et al.~\cite{fang2015captions} detected words from visual regions and used a maximum entropy language model to generate candidate captions. Instead of using LSTMs, they utilized a re-ranking method called deep multi-modal similarity to select the captions. Karpathy and Fei-Fei~\cite{karpathy2015deep} applied a region-based image captioning model consisting of two separate models to detect an image region and generate its corresponding caption. Johnson et al.~\cite{johnson2016densecap}, based on the work of Ren et al.~\cite{ren2015faster} on detecting image regions, incorporated the detection and generation tasks in an end-to-end training task. Attention mechanisms (either hard or soft) were applied by Xu et al.~\cite{xu2015show} to detect salient regions and generate their related words. In each time step, the model dynamically used the regional features as inputs to the LSTM model. The fragment-level encoding methods detect objects and their corresponding regions in an image. 
However, they usually neglect encoding fine and significant fragments of data such as emotions. 
The work that we describe next has recognised this: human captions, such as those in Fig.~\ref{fig:example1}, do include sentiment, and image captioning systems should therefore also aim to do this.

There are a few models that have incorporated sentiment into image captions \cite{gan2017stylenet,mathews2016senticap,you2018image}. However, this has typically required the construction of a new dataset, and the notion of sentiment is realized via a sentiment lexicon.
Mathews et al.~\cite{mathews2016senticap} applied a model to describe images using predefined positive and negative sentiments called SentiCap. The model used a full switching method including two parallel systems, each of which includes a Convolutional Neural Network (CNN) and a Long Short-Term Memory (LSTM). The first system was used to generate factual image captions and the second one to add word-level sentiments. The latter required a specifically constructed dataset, where crowdsourced workers rewrote thousands of factual captions to incorporate terms from a list of sentiment-bearing adjective-noun pairs. 
You et al.~\cite{you2018image} presented two optimum schemes to employ the predefined sentiments to generate image descriptions. Their approach is still focused on subjective descriptions of images using a given sentiment vocabulary, rather than representing the emotional content of the image.

Gan et al.~\cite{gan2017stylenet} StyleNet system that we noted in Sec.~\ref{sec:intro} adds styles, including sentiment values, to factual captions; these styles, such as humorous or romantic.  Once more, these reflect the attitude of the viewer to the image, and it is in principle possible to generate captions that do not accord with the content of the image: for instance, while happy faces of babies can be properly described using positive sentiment, it is difficult to apply negative sentiment in this context.

\noindent
In contrast to this work, we focus on images including human faces and recognize relevant emotions, using facial expression analyses, to generate image captions. Furthermore, we do not use any specific sentiment vocabulary or dataset to train our models: our goal is to see whether, given the existing vocabulary, incorporating facial emotion can produce better captions.

\subsection{Facial Expression Recognition}
Facial expression is a form of non-verbal communication which conveys attitude, affects, and intentions of individuals. Facial features and muscles changes during time lead to facial expression~\cite{fasel2003automatic}. Darwin started research leading to facial expressions more than one century ago~\cite{ekman2006darwin}. Now, there is a large body of work in recognizing basic facial expressions~\cite{fasel2003automatic,sariyanidi2015automatic} most often using the framework of six purportedly universal emotions \cite{ekman:1999} of happiness, sadness, fear, surprise, anger, and disgust plus neutral expressions. Recently, to find effective representations, deep learning based methods have been successfully applied to facial expression recognition (FER) tasks. 
%
%
They are able to capture hierarchical structures from low- to high-level data representations thanks to their complex architectures including multiple layers. Among deep models, Convolutional Neural Networks (CNNs) have achieved state-of-the-art performances in this domain. Kahou et al.~\cite{kahou2013combining}, as a winning submission to the 2013 Emotion Recognition in the Wild Challenge, used CNNs to recognize facial expressions. CNNs and linear support vector machines were trained to detect basic facial expressions by Tang~\cite{tang2013deep}, who won the 2013 FER challenge~\cite{goodfellow2013challenges}. In FER tasks, CNNs can be also used for transfer learning and feature extraction. Yu and Zhang~\cite{yu2015image} used CNNs, in addition to a face detection approach, to recognize facial expressions using transfer learning. The face detection approach was applied to detect faces areas and remove irrelevant noises in the target samples. Kahou et al.~\cite{kahou2016emonets} also used CNNs for extracting visual features together with audio features in a multi-modal framework.

As is apparent, these models usually employ CNNs with a fairly standard deep architecture to produce good results on the FER-2013 dataset~\cite{goodfellow2013challenges}, which is a large dataset collected `in the wild'. Pramerdorfer et al.~\cite{pramerdorfer2016facial}, instead, applied a combination of modern deep architectures including VGGnet~\cite{simonyan2014very} on the dataset. They succeeded in generating the state-of-the-art result in this domain. We similarly aim to train a facial expression recognition model that can recognize facial expressions in the wild and produce state-of-the-art performance on FER-2013 dataset. In the next step, we then use the model as a feature extractor on the images of FlickrFace11K, our extracted dataset from Flickr 30K~\cite{young2014image}. The features will be applied as a part of our image captioning models in this work.

\section{Describing an Image using Facial Expression Analysis}
\label{sec:models}

In this paper, we describe our image captioning models to generate image captions using facial expression analysis, which we term Face-Cap. We use a facial expression recognition model to extract the facial expression features from an image; the Face-Cap models in turn apply the features to generate image descriptions. In the following subsections, we first describe the datasets used in this work. 
Second, the face pre-processing step is explained to detect faces from our image caption data, and make them exactly similar to our facial expression recognition data. Third, the faces are fed into our facial expression recognition model to extract facial expression features.
%
%
%
Finally, we elucidate Face-Cap models, which are image captioning systems trained by leveraging additional facial expression features and image-caption paired data.

\subsection{Datasets}
\label{sec:datasets}

To train our facial expression recognition model, we use the facial expression recognition 2013 (FER-2013) dataset~\cite{goodfellow2013challenges}. It includes in-the-wild samples labeled \textit{happiness}, \textit{sadness}, \textit{fear}, \textit{surprise}, \textit{anger}, \textit{disgust}, and \textit{neutral}. It consists of 35,887 examples (28,709 for training, 3589 for public and 3589 for private test), collected by means of the Google search API. The examples are in grayscale at the size of 48-by-48 pixels. We split the training set of FER-2013 into two sections after removing 11 completely black examples: 25,109 for training and 3589 for validating the model. Similar to other work in this domain~\cite{kim2016fusing,pramerdorfer2016facial,yu2015image}, 
%
%
we use the private test set of FER-2013 for the performance evaluation of the model after the training phase. To compare with the related work, we do not apply the public test set either for training or for validating the model.
%

To train our image captioning models, we have extracted 
a subset of the Flickr 30K dataset with image captions~\cite{young2014image}, which we term
FlickrFace11K. It contains 11,696
%
%
examples including human faces, which are detected using a CNN-based face detection algorithm~\cite{king2009dlib}.\footnote{The new version (2018) of Dlib library is applied.} We observe that the Flickr 30K dataset is a good source for our dataset, because it has a larger portion of samples that include human faces, in comparison with other image caption datasets such as the COCO dataset~\cite{chen2015microsoft}. We split the FlickrFace11K samples into 8696 for training, 2000 for validation and 1000 for testing, and make them publicly available.\footnote{\url{https://github.com/omidmn/Face-Cap}} To extract the facial features of the samples, we use a face pre-processing step and a facial expression recognition model as follows.

\subsection{Face Pre-processing}
\label{sec:model-face-regis}
Since we aim to train a facial expression recognition model on FER-2013 and use it as a facial expression feature extractor on the samples of FlickrFace11K, we need to make the samples consistent with the FER-2013 data. To this end, a face detector is used to pre-process the faces of FlickrFace11K. The faces are detected by the CNN-based face detection algorithm and cropped from each sample. Then, we transform each face to grayscale and resize it into 48-by-48 pixels, which is exactly the same FER-2013 data.

\subsection{Facial Expression Recognition Model}
\label{sec:model-fer}
%
%
In this section, using the FER-2013 dataset, we train a VGGnet model~\cite{simonyan2014very} to recognize facial expressions. The model's architecture is similar to recent work~\cite{pramerdorfer2016facial} that is state-of-the-art in this domain, and our replication gives similar performance. The classification accuracy, which is a popular performance metric on the FER-2013 dataset,
%
%
on the test set of FER-2013 is $72.7\%$. It is around $7\%$ better than the human performance ($65\pm5\%$) on the test set~\cite{goodfellow2013challenges}. The output layer of the model, generated using a softmax function, includes seven neurons, corresponding to the categorical distribution probabilities over the emotion classes in FER-2013 including \textit{happiness}, \textit{sadness}, \textit{fear}, \textit{surprise}, \textit{anger}, \textit{disgust}, and \textit{neutral}; we refer to this by the vector $a = (a_1, \ldots, a_7)$.

We use the network to extract the probabilities of each emotion from all faces, as detected in the pre-processing step of Sec.~\ref{sec:model-face-regis}, in each FlickrFace11K sample. 




For each image, we construct a vector of facial emotion features $s = (s_1, \ldots, s_7)$ used in the Face-Cap models as in Eq.~\ref{eq:example11}.

\begin{equation}
s_k = \begin{cases}
1\ \ \ \textnormal{for}\   k=\arg \max \sum_{1 \leq i \leq n}{a_{i}},\\
0\ \ \ \textnormal{otherwise}
\end{cases}
\label{eq:example11}
\end{equation}

\noindent
where $n$ is the number of faces in the sample. That is, $s$ is a one-hot encoding of the aggregate facial emotion features of the image.

\subsection{Training Face-Cap}

\subsubsection{Face-Cap$_F$}
In order to train the Face-Cap models, we apply a long short-term memory (LSTM) network as our caption generator, adapted from Xu et al.~\cite{xu2015show}. The LSTM is informed about the emotional content of the image using the facial features, defined in Eq.~\ref{eq:example11}.
It also takes the image features which are extracted by Oxford VGGnet~\cite{simonyan2014very}, learned on the ImageNet dataset, and weighted using the attention mechanism~\cite{xu2015show}. In the mechanism, the attention-based features, including the factual content of the image, are chosen for each generated word in the LSTM.
Using Eq.~\ref{eq:example2}, in each time step ($t$), the LSTM uses the previously embedded word ($x_{t-1}$), the previous hidden state ($h_{t-1}$), the image features ($z_{t}$), and the facial features ($s$) to generate input gate ($i_t$), forget gate ($f_t$), output gate ($o_t$), input modulation gate ($g_t$), memory cell ($c_t$), and hidden state ($h_t$).

\begin{equation}
\begin{split}
& i_t = \sigma(W_{i}x_{t-1} + U_{i}h_{t-1} + Z_{i}z_{t} + S_{i}s + b_i) \\
& f_t = \sigma(W_{f}x_{t-1} + U_{f}h_{t-1} + Z_{f}z_{t} + S_{f}s + b_f) \\
& o_t = \sigma(W_{o}x_{t-1} + U_{o}h_{t-1} + Z_{o}z_{t} + S_{o}s + b_o) \\
& g_t = \tanh(W_{c}x_{t-1} + U_{c}h_{t-1} + Z_{c}z_{t} + S_{c}s + b_c) \\
& c_t = f_{t}c_{t-1}+i_{t}g_t \\
& h_t = o_t\tanh(c_t) \quad
\end{split}
\label{eq:example2}
\end{equation} where $W, U, Z, S$, and $b$ are learned weights and biases and $\sigma$ is the logistic sigmoid activation function. According to Eq.~\ref{eq:example2}, the facial features of each image are fixed in all time steps and the LSTM automatically learns to condition, at the appropriate time, the next generated word by applying the features.
%
To initialize the LSTM's memory state ($c_0$) and hidden state ($h_0$), we feed the facial features through two typical multilayer perceptrons, shown in Eq.~\ref{eq:example22}.

\begin{equation}
c_0=\tanh_{init,c}( s ), \ \ h_0=\tanh_{init,h}( s )
\label{eq:example22}
\end{equation}

\noindent
We use the current hidden state ($h_t$), to calculate the negative log-likelihood of $s$ in each time step (Eq.~\ref{eq:example3}), 
named the face loss function.
Using this method, $h_t$ will be able to record a combination of $s$, $x_{t-1}$ and $z_{t}$ in each time step.

\begin{equation}
L(h_t, s)=-\sum_{1 \leq i \leq 7}1_{(i=s)}\log(p(i|h_t)) \quad
\label{eq:example3}
\end{equation} where a multilayer perceptron generates $p(i|h_t)$, which is the categorical probability distribution of the current state across the emotion classes. 
In this we adapt You et al.~\cite{you2018image}, who use this loss function for injecting ternary sentiment (positive, neutral, negative) into captions.
This loss is estimated and averaged, over all steps, during the training phase.

\subsubsection{Face-Cap$_L$}
\begin{figure}
	\centering
	\includegraphics[width=\textwidth]{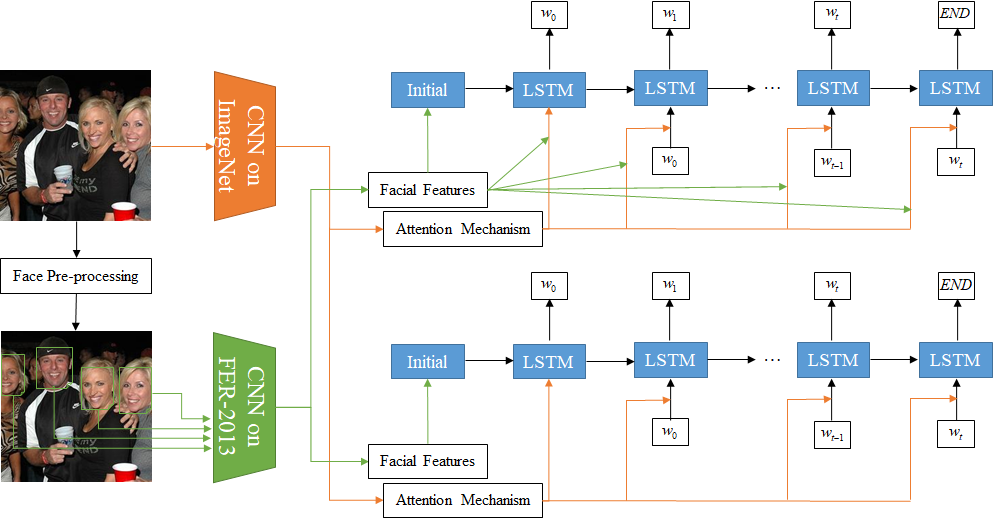}
	\caption{
		The frameworks of Face-Cap$_{F}$ (top), and Face-Cap$_{L}$ (bottom). The face pre-processing and the feature extraction from the faces and the image are illustrated. The Face-Cap models are trained using the caption data plus its corresponding image features, selected using the attention mechanism, and facial features.
	}
	\label{fig:example55}
\end{figure}

The above Face-Cap$_F$ model feeds in the facial features at the initial step (Eq.~\ref{eq:example22}) and at each time step (Eq.~\ref{eq:example2}), shown in Fig.~\ref{fig:example55} (top). In Face-Cap$_{F}$, the LSTM uses the facial features for generating every word because the features are fed at each time step. Since a few words, in the ground truth captions (e.g. Fig.~\ref{fig:example1}), are related to the features, this mechanism can sometimes lead to less effective results.

Our second variant of the model, Face-Cap$_L$, is as above except that the $s$ term is removed from Eq.~\ref{eq:example2}: we do not apply the facial feature information at each time step (Fig.~\ref{fig:example55} (bottom)), eliminating it from Eq.~\ref{eq:example2}. Using this mechanism, the LSTM can effectively take the facial features in generating image captions and ignore the features when they are irrelevant. To handle this issue, You et al.~\cite{you2018image} implemented the sentiment cell, working similar to the memory cell in the LSTM, initialized by the ternary sentiment. They fed the image features to initialize the memory cell and hidden state of the LSTM.
In comparison, Face-Cap$_L$ uses the facial features to initialize the memory cell and hidden state rather than the sentiment cell which requires more time and memory to compute. Using the attention mechanism, our model applies the image features in generating every caption word.

\section{Experiments}
\label{sec:exper}

\subsection{Evaluation Metrics and Testing} To evaluate Face-Cap$_{F}$ and Face-Cap$_{L}$, we use standard evaluation metrics including BLEU~\cite{papineni2002bleu}, ROUGE-L~\cite{lin2004rouge}, METEOR~\cite{denkowski2014meteor}, CIDEr~\cite{vedantam2015cider}, and SPICE~\cite{anderson2016spice}. All five metrics with larger values mean better results.

We train and evaluate all models on the same splits of FlickrFace11K.

\subsection{Models for Comparison} The model of Xu et al.~\cite{xu2015show} is the starting point of Face-Cap$_{F}$ and Face-Cap$_{L}$, which is selectively attending to a visual section at each time step. We train Xu's model using the FlickrFace11K dataset. 

We also look at two additional models to investigate the impact of the face loss function in using the facial features in different schemes.
We train the Face-Cap$_{F}$ model, which uses the facial features in every time step, without calculating the face loss function (Eq.~\ref{eq:example3}); we refer to this as the Face-Step model. The Face-Cap$_{L}$ model, which applies the facial features in the initial time step, is also modified in the same way; we refer to this as the Face-Init model.

\subsection{Implementation Details} In our implementation, the memory cell and the hidden state of the LSTM each have 512 dimensions.\footnote{We use TensorFlow to implement the models~\cite{abadi2016tensorflow}.} We set the size of the word embedding layer to 300, which is initialized using a uniform distribution. The mini-batch size is 100 and the epoch limit is 20. We train the models using the Adam optimization algorithm~\cite{kingma2014adam}. The learning rate is initialized to 0.001, while its minimum is set to 0.0001. If there is no improvement of METEOR for two successive epochs, the learning rate is divided by two and the prior network that has the best METEOR is reloaded.
%
%
This approach leads to effective results in this work. For Adam, tuning the learning rate decay, similar to our work, is supported by Wilson et al.~\cite{wilson2017marginal}. METEOR on the validation set is used for model selection.
%
We apply METEOR for the learning rate decay and the model selection because it shows reasonable correlation with human judgments but calculates more quickly than SPICE (as it does not require dependency parsing) \cite{anderson2016spice}.

%
%
Exactly the same visual feature size and vocabulary are used for all five models. As the encoder of images, in this work as for Xu et al., we use Oxford VGGnet~\cite{simonyan2014very} trained on ImageNet, and take its fourth convolutional layer (after ReLU), which gives $14\times14\times512$ features. 
For all five models, the negative log likelihood of the generated word is calculated, as the general loss function, at each time step.
\begin{table}
	\caption{Comparisons of image caption results (\%) on the test split of FlickrFace11K dataset. B-1, ... SPICE are standard evaluation metrics, where B-N is BLEU-N metric.}
	\label{tab:metrics}
	\begin{center}
		\begin{tabular}{|l|l|l|l|l|l|l|l|l|}
			\hline
			Model & B-1 & B-2 & B-3 & B-4 & METEOR & ROUGE-L & CIDEr & SPICE\\
			\hline
			Xu's model & 55.95  & 35.43 & 23.06 & 15.69 & 16.96 & 43.71 & 21.94 & 9.30 \\
			Face-Step & 58.43  & 37.56 & 24.78 & 16.96 & \textbf{17.45} & 45.04 & 22.83 & 9.90 \\
			Face-Init & 56.63  & 36.49 & 24.30 & 16.86 & 17.17 & 44.84 & 23.13 & 9.80 \\
			Face-Cap$_{F}$ & 57.13 & 36.51 & 24.07 & 16.52 & 17.19 & 44.76 & 23.04 & 9.70 \\
			Face-Cap$_{L}$ & \textbf{58.90} & \textbf{37.89} & \textbf{25.07} & \textbf{17.19} & \textbf{17.44} & \textbf{45.47} & \textbf{24.72} & \textbf{10.00}  \\
			\hline
			
		\end{tabular}
	\end{center}
\end{table}

\subsection{Results} 
\label{sec:results}

\subsubsection{Overall Metrics}
The experimental results are summarized in Table~\ref{tab:metrics}.  All Face models outperform Xu's model using all standard evaluation metrics. This shows that the facial features are effective in image captioning tasks. As predicted, Face-Cap$_{L}$ has a better performance in comparison with other models using all the metrics except METEOR, where it is only very marginally (0.01) lower. Under most metrics, Face-Step performs second best, with the notable exception of CIDEr, suggesting that its strength on other metrics might be from use of popular words (which are discounted under CIDEr).  Comparing the mechanics of the top two approaches, Face-Cap$_{L}$ uses the face loss function to keep the facial features and apply them at the appropriate time; however, Face-Step does not apply the face loss function. Face-Cap$_{L}$ only applies the facial features in the initial time step, while Face-Step uses the features in each time step, in generating an image caption. In this way, Face-Step can keep the features without applying the face loss function. This yields comparable results between Face-Cap$_{L}$ and Face-Step; however, the results show that applying the face loss function is more effective than the facial features in each time step. This relationship can also be seen in the results of Face-Init, which is Face-Cap$_{L}$ without the face loss function. The results of Face-Cap$_{F}$ show that a combination of applying the face loss function and the facial features in each time step is problematic.

\begin{table}
	\caption{Comparisons of distributions of verbs in generated captions: entropies, and probability mass of the top 4 frequent verbs (\textit{is}, \textit{sitting}, \textit{are}, \textit{standing})}
	\label{tab:verb-distrib}
	\begin{center}
		\begin{tabular}{|l|l|l|}
			\hline
			Model & Entropy & Top 4\\
			\hline
			Xu's model & 2.7864 & 77.05\%\\
			Face-Step & 2.9059 & 74.80\% \\
			Face-Init & 2.6792 & 78.78\%\\
			Face-Cap$_{F}$ & 2.7592 & 77.68\%\\
			Face-Cap$_{L}$ & \textbf{2.9306} & \textbf{73.65\%} \\
			\hline
		\end{tabular}
	\end{center}
\end{table}

\begin{table}
	\caption{The ranks of sample generated verbs under each model.}
	\label{tab:verb-examples}
	\begin{center}
		\begin{tabular}{|l|l|l|l|l|l|l|l|l|}
			\hline
			Model & Smiling & Looking & Singing & Reading & Eating & Laughing \\
			\hline
			Xu's model & 19 & n/a & 15 & n/a & 24 & n/a \\
			Face-Step & 11 & \textbf{18} & 10 & n/a & 15 & n/a\\
			Face-Init & 10 & 21 & 12 & n/a & 14 & n/a\\
			Face-Cap$_{F}$ & 12 & 20 & \textbf{9} & n/a & 14 & n/a \\
			Face-Cap$_{L}$ & \textbf{9} & \textbf{18} & 15 & \textbf{22} & \textbf{13} & \textbf{27} \\
			\hline
		\end{tabular}
	\end{center}
\end{table}

\subsubsection{Caption Analysis}
To analyze what it is about the captions themselves that differs under the various models, with respect to our aim of injecting information about emotional states of the faces in images, we first extracted all generated adjectives, which are tagged using the Stanford part-of-speech tagger software~\cite{toutanova2003feature}.  Perhaps surprisingly, emotions do not manifest themselves in the adjectives in Face-Cap models: the adjectives used by all systems are essentially the same. This may be because adjectives with weak sentiment values (e.g. \textit{long}, \textit{small}) predominate in the training captions, relative to the adjectives with strong sentiment values (e.g. \textit{happy}, \textit{surprised}).

We therefore also investigated the difference in distributions of the generated verbs under the models. Entropy (in the information-theoretic sense) can indicate which distributions are closer to deterministic and which are more spread out (with a higher score indicating more spread out) calculated using Eq.~\ref{eq:example17}.

\begin{equation}
E = -\sum_{1 \leq i \leq n}{p(x_i)\times\log_2(p(x_i))} \quad
\label{eq:example17}
\end{equation} where $E$ is the entropy score and $n$ is the number of the generated unique verbs under each model. $P(x_i)$ is the probability of each generated unique verb ($x_i$), estimated as the Maximum Likelihood Estimate from the sample. From Table~\ref{tab:verb-distrib}, Face-Cap$_{L}$ has the highest entropies, or the one with the greatest variability of expression. Relatedly, we look at the four most frequent verbs, which are the same for all models (\textit{is}, \textit{sitting}, \textit{are}, \textit{standing}) --- these are verbs with relatively little semantic content, and for the most part act as syntactic props for the content words of the sentence. Table~\ref{tab:verb-distrib} also shows that Face-Cap$_{L}$ has the lowest proportion of the probability mass taken up by these, leaving more for other verbs.

\begin{figure}
	\centering
	\includegraphics[height=13.1 cm]{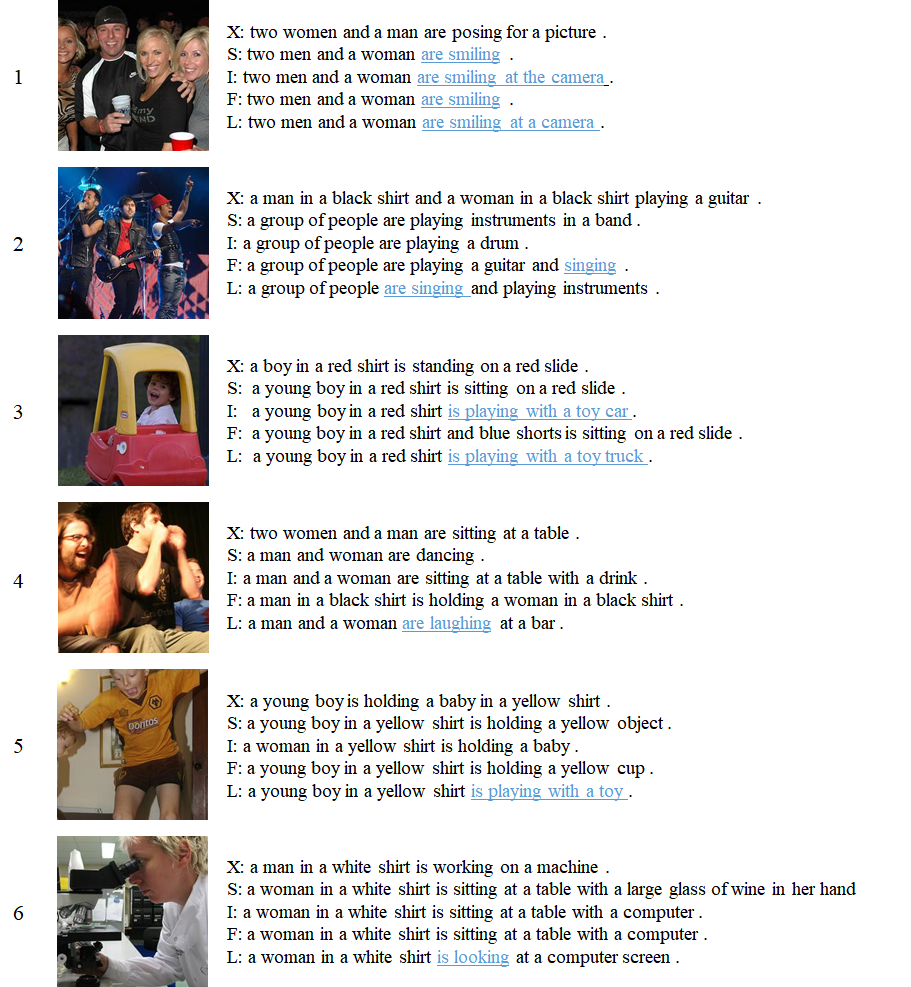}
	\caption{Examples of different image captioning models including X (Xu's model), S (Face-Step), I (Face-Init), F (Face-Cap$_{F}$), and L (Face-Cap$_{L}$).}
	\label{fig:example3}
\end{figure}

The ranks of the generated verbs under the models, which are calculated using the numerical values of their frequency, are also interesting. Table~\ref{tab:verb-examples} includes some example verbs; of these, \textit{smiling}, \textit{singing}, and \textit{eating} are higher ranked under the Face-Cap models, and \textit{reading} and \textit{laughing} only appear under the Face-Cap$_{L}$ model. \textit{Looking} is also generated only using the models including the facial features. These kinds of verbs are relevant to the facial features and show the effectiveness of applying the features in generating image captions.

\begin{figure}
	\centering
	\includegraphics[height=6.8 cm]{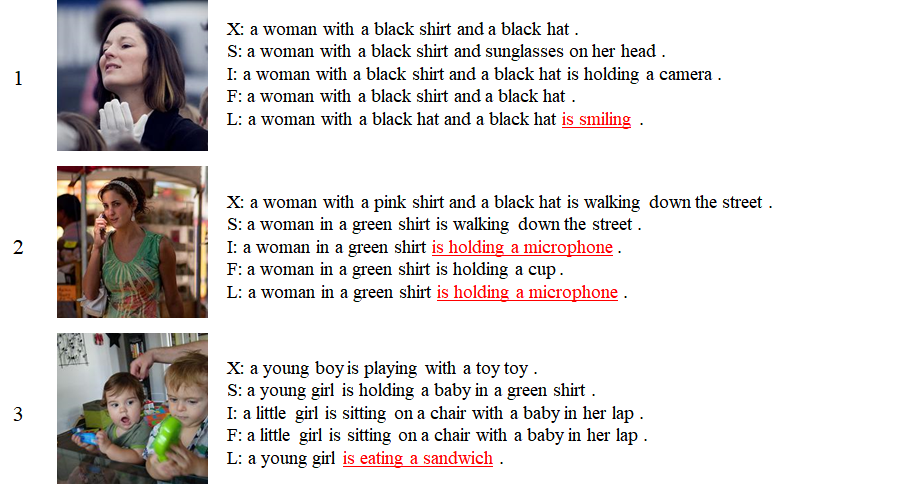}
	\caption{Examples of the models including various amounts of error.}
	\label{fig:example4}
\end{figure}

\subsubsection{Samples}
Fig.~\ref{fig:example3} includes a number of generated captions, for six sample images, under all models in this work. In example 1, the models that include facial features properly describe the emotional content of the image using \textit{smiling}. The Face-Cap$_{L}$ model also generates \textit{laughing} according to the emotional content of example 4. In example 3, the Face-Init and the Face-Cap$_{L}$ models generate \textit{playing} which is connected to the emotional content of the example. It is perhaps because the child in the example is happy that the models generate \textit{playing}, which has a positive sentiment connotation. In example 5, Face-Cap$_{L}$ also uses \textit{playing} in a similar way. Example 2 shows that the Face-Cap models apply \textit{singing} at the appropriate time. Similarly, \textit{looking} is used, by Face-Cap$_{L}$, in example 6. \textit{Singing} and \textit{looking} are generated because of the facial features of people in the examples, which are related to some emotional states such as \textit{surprised} and \textit{neutral}. Fig.~\ref{fig:example3} shows that our models can effectively apply the facial features to describe images in different ways. In Fig.~\ref{fig:example4}, three examples are shown, which our models inappropriately use the facial features. \textit{Smiling} is used to describe the emotional content of the example 1; however, the girl in the example is not happy. The results of the example 2 and 3 wrongly contain \textit{holding a microphone} and \textit{eating}, which are detected from the facial features, due to visual likeness.

\section{ Conclusion and Future Work}
\label{sec:concl}

In this paper, we have proposed two variants of an image captioning model, Face-Cap, which employ facial features to describe images. To this end, a facial expression recognition model has been applied to extract the features from images including human faces. Using the features, our models are informed about the emotional content of the images to automatically condition the generating of image captions. We have shown the effectiveness of the models using standard evaluation metrics compared to the state-of-the-art baseline model. The generated captions demonstrate that the Face-Cap models succeed in generating image captions, incorporating the facial features at the appropriate time. Linguistic analyses of the captions suggest that the improved effectiveness in describing image content comes through greater variability of expression. 

Future work can involve designing new facial expression recognition models, which can cover a richer set of emotions including \textit{confusion} and \textit{curiousity}; and effectively apply their corresponding facial features to generate image captions. In addition, we would like to explore alternative architectures for injecting facial emotions, like the soft injection approach of \cite{xu2015show}.

%
%
%
%


\end{document}